\documentclass{article}
\usepackage{spconf}
\usepackage{cite}
\usepackage{amsmath,amssymb,amsfonts}
\usepackage{algorithmic}
\usepackage{graphicx}
\usepackage{textcomp}
\usepackage{xcolor}
\usepackage{booktabs}
\usepackage{tabularx} 
\usepackage{placeins} 
\usepackage[hidelinks]{hyperref} 
\usepackage{url} 

\newcolumntype{Y}{>{\hsize=0.75\hsize\raggedright\arraybackslash}X}

\begin{document}

\title{Training Intelligent Voice Assistant Wakeup with Controllable Synthetic Conversations}

\twoauthors
  {Marcin Sowa\'nski, Kacper Leszczy\'nski}
    {TCL Research Europe\\
     \texttt{marcin.sowanski@tcl.com}\\
     \texttt{kacper.leszczynski@tcl.com}}
  {Kacper Krzywicki, Krzysztof Wodnicki}
    {University of Warsaw\\
     \texttt{k.krzywicki@student.uw.edu.pl}\\
     \texttt{k.wodnicki@student.uw.edu.pl}}

\maketitle

\begin{abstract}
Wake word detection is a critical component of virtual assistants, serving as the gateway to seamless user interactions. This paper introduces a novel wake-up system that extends traditional direct keyword detection with contextual trigger detection. After an initial wake word activation, the system uses reasoning to distinguish between user commands and unrelated speech, ensuring efficient and context-aware engagement. We present a data generation architecture that produces a 62.3-hour corpus of controllable multi-speaker conversations containing direct invocations, contextual follow-ups, and non-addressed speech. Experimental results demonstrate the effectiveness of the proposed approach across diverse synthetic conversational scenarios. We release the code, dataset and trained models to promote reproducibility and further advancements in intelligent assistant technologies.\footnote{\href{https://tclresearcheurope.github.io/intelligent_wakeup/}{\url{https://tclresearcheurope.github.io/intelligent_wakeup/}}}
\end{abstract}


\section{Introduction}

Virtual assistants are becoming more and more present in daily life, but their interactions remain largely constrained by turn-based, command-driven paradigms. Traditionally, the wake-up mechanism relies on a specific key word or phrase, typically the assistant's name, and only after detecting the phrase does it trigger a more powerful model, such as a Large Language Model (LLM) that is responsible for handling the actual response. Although computationally efficient, this keyword-dependent architecture interrupts the natural flow of conversation and forces users into rigid interaction patterns.

These limitations motivate a novel intelligent wake-up system that determines whether to activate an assistant based on the current utterance and conversational context. This formulation changes the role of the wake-up component from a simple keyword detector into a lightweight decision-making module serving as a gateway to more complex user interactions.

\begin{figure}[!t]
    \centering
    \includegraphics[scale=0.8]{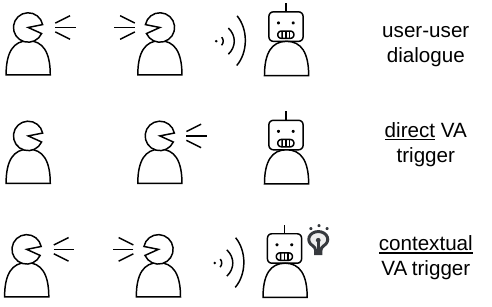}
    \caption{Overview of the intelligent wakeup usage scenario.}
    \label{fig:iw-usecase-diagram}
\end{figure}

In this work, we propose a speech classification model that combines an initial wake-up mechanism with contextual trigger detection to determine whether subsequent utterances require an assistant response. As illustrated in Fig.~\ref{fig:iw-usecase-diagram}, the system distinguishes user--user dialogue, direct assistant activation, and contextual activation. Finally, we develop a novel framework for generating synthetic multi-speaker conversational scenarios and release the corresponding code and datasets to promote reproducibility and further research.

\section{Related Work}

\begin{figure*}[!t]
    \centering
    \includegraphics[width=\textwidth]{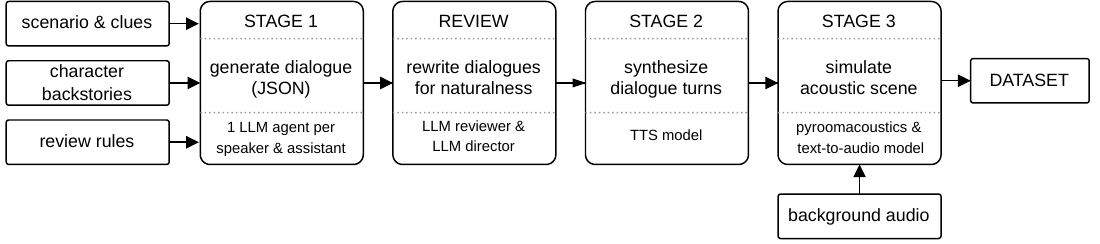}
    \caption{Overview of the synthetic conversation generation pipeline.}
    \label{fig:iw-text_gen_pipeline-diagram}
\end{figure*}

Historically, research in wake-up models has focused heavily on optimizing streaming keyword spotting for resource-constrained edge devices\cite{lvarez2018EndtoendSK}. Standard approaches utilize end-to-end streaming networks and streaming transformers to achieve low-latency recognition. Several systems emphasize highly accurate on-device performance utilizing background modeling and multi-task learning for endpoint detection\cite{tang2020howldeployedopensourcewake,Ramanan2020RecipeFC,Wu2018monophone,maekaku19_interspeech}. Although highly optimized, these systems often struggle with varied acoustic environments and are fundamentally limited to pre-defined key phrases, which require specialized techniques such as prototypical networks to adapt to new words\cite{Parnami_2022}.

Moving beyond static keywords, recent literature highlights the need to use contextual cues for activation. Research has explored incorporating environmental and speaker-specific information to personalize key phrase detection\cite{rikhye21_interspeech}.

The goal of eliminating wake words is to enable seamless human-like interaction. This has led to the development of full-duplex dialogue architectures that break the traditional turn-taking approach\cite{Lin_2022}. Recent breakthroughs include models capable of listening while simultaneously speaking\cite{ma2025language}, real-time speech-text foundation models like Moshi\cite{defossez2024moshispeechtextfoundationmodel}, and various full-duplex schemes built upon LLMs\cite{wang2024full,fang2025llama}.


The work most closely related to ours is Cocktail-Talker \cite{jiang2026cocktail}, which studies selective participation of a speech agent in a noisy multi-speaker environment. Cocktail-Talker models three possible agent actions: \textit{respond}, \textit{listen}, and \textit{ignore}. While our work shares the related goal of determining whether an assistant should respond, we focus on a different type of activation; Cocktail-Talker decides if an agent should participate in each turn, whereas our system determines if the assistant should continue the interaction without requiring another wake word.

\section{Data and Method}

We define intelligent wakeup as a turn-level speech classification task. Given the current utterance and the preceding conversation history, the model determines whether the virtual assistant should be activated. In contrast to traditional wake word detection, activation may be direct, through an explicit wake word, or contextual, when a follow-up utterance implicitly addresses the assistant based on previous turn.

In this study, we propose a novel framework for detecting interactions with virtual assistants. The proposed method comprises two key components: (1) data generation; and (2) intelligent wakeup detection model.

\subsection{Data Generation}

Our dataset is designed to cover the task space of the formal wake-up behavior model. We developed a tool for generating synthetic, coherent, and natural conversations for this setting consisting of three phases: text generation, speech synthesis for individual dialogue turns, and audio composition in a simulated environment, as illustrated in Fig.~\ref{fig:iw-text_gen_pipeline-diagram}.

\subsubsection{Scenario Specification}

Each scenario is a declarative specification of the structure of conversation defined by a sequence of phases. Each phase names participating users in turn order, a topic and mood. The assistant participates only in the phases that list it among the speakers. To keep conversations within our requirements, we also applied deterministic post-editing rules.

\subsubsection{Text Data}

We used one LLM agent per human speaker and a separate assistant agent, with speaker-specific prompt and character backstories describing personality and relationships. Reviewer and director stages further rewrote generated dialogues for naturalness and coherence.

\subsubsection{Speech Data}

We synthesized each dialogue turn independently with ElevenLabs TTS; however, we supported open-source models: Qwen-TTS, Kokoro, Bark, MMS, SpeechT5, ParlerTTS, and Narakeet. 

\subsubsection{Background Sounds}

The final audio files consist of background sounds generated by either a text-to-audio model (ElevenLabs, Kokoro) or sounds from the Freesound platform (data with Creative Commons 0 license).

\subsection{Dataset}

The first version of the dataset (v1.0.1)\footnote{\href{https://huggingface.co/datasets/TCLResearchEurope/intelligent_wakeup}{\url{https://huggingface.co/datasets/TCLResearchEurope/intelligent_wakeup}}} consists of 1001 core scenarios grouped into 59 categories that represent real-life situations where a virtual assistant can be used. Our generation framework generated 2,297 conversations, 35,788 dialogue turns, and 62.30 h of audio that was then split into train (50.15h), valid (5.43h) and test (6.72h) parts. 85.7\% of conversations have two or more human speakers (other than assistant). To prevent distribution bias, the voice assistant is present in 66\% of all scenarios, and the remaining scenarios contain conversations among two or more speakers without assistant involvement.

\section{Experiments}

\subsection{TTS Speakers Voice Analysis}

We evaluated 169 unique speakers generated with ElevenLabs voice design model against VCTK v0.92 \cite{vctk2013} and two Qwen3-TTS baselines (LibriTTS-R and speakers designed from text prompt). We used eight distinct utterances per speaker, creating 4,376 utterances. The eight utterances setting provides 28 intra-speaker pairs per speaker and makes all systems directly comparable.

We embedded the speaker utterances with ECAPA-TDNN (192-d)\cite{desplanques20_interspeech} and WavLM-SV (512-d)\cite{chen2022wavlm}. Using L2-normalized embeddings and cosine distance, we measured intra-speaker distance, inter-speaker centroid distance, and equal-error rate (EER).

The selected encoders differ substantially in input representation, architecture family, training paradigm and pre-training data (neither encoder has seen synthetic speech during training), reducing the dependence on a single representation. Both are trained discriminatively on speaker identity, so they measure whether a speaker-recognition system would treat two clips as the same person.

\subsection{Corpus Naturalness Evaluation}

We evaluated naturalness of the proposed intelligent wakeup dataset against natural conversational audio from NOTSOFAR-1. Our primary objective was to test whether it is non-inferior in conversational naturalness using a five-point rating scale. In addition, we tested interaction naturalness, scenario plausibility and conversational coherence to find potential differences between the corpora.

We sampled 30 clips from each corpus, each lasting 1--3 minutes. We recruited 120 annotators through Prolific and compensated them at a rate of £18 per hour. Each annotator rated up to eight clips, balanced across the two corpora. After applying quality control, 900 valid ratings remained, corresponding to 15 ratings per clip.

For each question, we fit a linear mixed-effects model with corpus as the main fixed effect and crossed random intercepts for annotator and clip; scenario and clip duration are included as additional fixed effects. Non-inferiority is established when the lower bound of the one-sided 95\% confidence interval for the proposed-minus-natural corpus difference exceeds -0.30. We used a cumulative-link mixed model as a robustness analysis to account explicitly for the ordinal nature of the ratings. Under approximate variance assumptions, the design provides about 83\% power to establish non-inferiority when the true corpus difference is zero.

\subsection{Intelligent Wakeup}

The task is offline and turn-level: the input is one completed turn plus encoded history, and the wake decision is made once after the turn ends, not incrementally.

As a reference, we prompted \texttt{gpt-realtime-mini} on the test split of our dataset, one real-time session per conversation so context accumulated. We disabled server VAD and used turn boundaries from the corpus so both systems saw identical segmentation.

All trained models\footnote{\href{https://huggingface.co/collections/TCLResearchEurope/intelligent-wakeup-models}{\url{https://huggingface.co/collections/TCLResearchEurope/intelligent-wakeup-models}}} share one architecture in which only the per-turn encoder varies. A turn is passed through a pre-trained speech encoder and its frame sequence collapsed to one vector by attention pooling. Every utterance is truncated to the 30 s input window imposed by Whisper's feature extractor. The pooled vector is concatenated with a learned speaker embedding, projected to 256 dimensions, given a positional embedding and passed to a causal transformer over the turn sequence: four pre-norm layers, four heads, up to twenty preceding turns, causally masked. Two heads read the contextualized current turn. This stack is 3.4 M parameters regardless of encoder.

The test set contains 158 direct, 141 contextual and 3,348 non-addressed turns. Direct and contextual recall are reported separately, while F1 denotes the positive-class F1 after combining both activation types. Predictions are positive when the activation probability exceeds a per-model threshold. All hyperparameters and the decision threshold were selected exclusively on the validation set.

\section{Results}

\subsection{TTS Speakers Voice Analysis}

\begin{figure}[ht]
    \centering
    \includegraphics[scale=0.45]{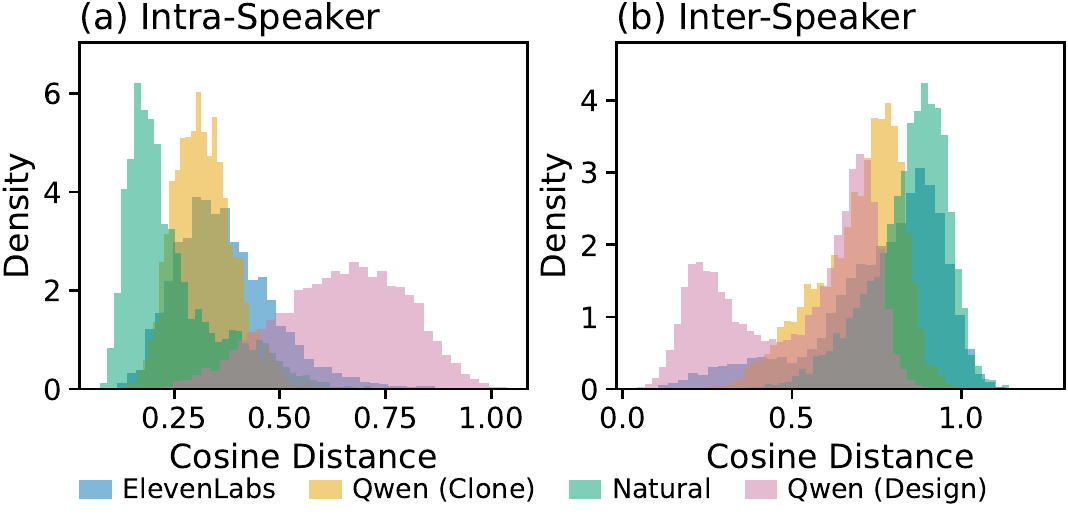}
    \caption{ECAPA-TDNN cosine distances: (a) intra-speaker, over utterance pairs, and (b) inter-speakers, over speaker centroids.}
    \label{fig:iw-tts_voice-ecapa-diagram}
\end{figure}

Across both encoders every system produces intra-speaker distances well below its inter-speaker distances (Table ~\ref{tab:iw-tts_analysis-distances}, Fig. ~\ref{fig:iw-tts_voice-ecapa-diagram}). VCTK separates best under ECAPA-TDNN, followed by Qwen cloning and ElevenLabs. Under WavLM-SV, Qwen cloning outperforms VCTK, likely because synthesis removes recording and session variability.

Reference-free Qwen voice design behaves differently: intra-speaker distance approaches inter-speaker distance, with $\approx$28\% EER under both encoders. Its intra- and inter-speaker distributions strongly overlap, including clusters of nominally different voices mapped very close together.

\begin{table}[t]
    \centering
    \footnotesize
    \begin{tabular}{lccc|ccc}
        \toprule
        & \multicolumn{3}{c}{\textit{ECAPA-TDNN}}
        & \multicolumn{3}{c}{\textit{WavLM-SV}} \\
        System & Intra & Inter & EER
               & Intra & Inter & EER \\
        \midrule
        ElevenLabs    & .372 & .838 & 6.2\%
                      & .075 & .370 & 11.2\% \\
        Qwen (Clone)  & .321 & .783 & 1.3\%
                      & .052 & .318 & 5.7\% \\
        Qwen (Design) & .644 & .792 & 28.6\%
                      & .139 & .389 & 27.7\% \\
        Natural       & .246 & .878 & 1.0\%
                      & .053 & .324 & 8.5\% \\
        \bottomrule
    \end{tabular}
    \caption{Intra-/inter-speaker cosine distance and EER. Distances are encoder-specific.}
    \label{tab:iw-tts_analysis-distances}
\end{table}

Despite different absolute EERs, both encoders preserve the same overall ordering, indicating that the effect reflects the synthesis method rather than a particular embedding space.

\subsection{Corpus Naturalness Evaluation}

Table \ref{tab:iw-corpus_naturalness_statistics} shows aggregated mean, median and standard deviation across all ratings for both datasets. The non-inferiority criterion was not met for any evaluated dimension. Our proposed dataset scores lower on all metrics, indicating higher perceived
naturalness of the NOTSOFAR-1 dataset. This is further supported by the results in Table \ref{tab:iw-corpus_naturalness} where the estimated effect for each dimension is negative. However, the disparity in the results for the \textit{scenario plausibility} and \textit{conversational coherence} is noticeably smaller than for other dimensions. This suggests that while generated conversations are quite plausible and rather coherent, the biggest disparity lies in more nuanced speech generation that would be able to express human emotions and react to other speakers naturally.

\begin{table}[t]
    \centering
    \small
    \setlength{\tabcolsep}{2.5pt}
    \begin{tabular}{lcccccc}
        \hline
        \textbf{Dimension}
        & \multicolumn{2}{c}{\textbf{Mean}}
        & \multicolumn{2}{c}{\textbf{Median}}
        & \multicolumn{2}{c}{\textbf{Std}} \\
        \cline{2-7}
        & \textbf{Nat.} & \textbf{Ours}
        & \textbf{Nat.} & \textbf{Ours}
        & \textbf{Nat.} & \textbf{Ours} \\
        \hline
        Conversation & 4.16 & 2.69 & 4.0 & 2.0 & 1.04 & 1.16 \\
        Speech       & 4.26 & 2.77 & 4.0 & 2.0 & 0.92 & 1.19 \\
        Interaction  & 4.13 & 2.81 & 4.0 & 3.0 & 1.07 & 1.17 \\
        Scenario     & 4.18 & 3.65 & 4.0 & 4.0 & 0.92 & 1.05 \\
        Coherence    & 4.07 & 3.64 & 4.0 & 4.0 & 0.99 & 1.02 \\
        \hline
    \end{tabular}
    \caption{Naturalness ratings for natural and synthetic conversations.}
    \label{tab:iw-corpus_naturalness_statistics}
\end{table}

\begin{table}[ht]
    \centering
    \begin{tabular}{lrrr}
        \hline
        \textbf{Dimension} & \textbf{EE} & \textbf{SE} & \textbf{LB} \\
        \hline
        \textbf{Overall conv. naturalness} & \textbf{-1.47} & \textbf{0.095} & \textbf{-1.63} \\
        \hline\hline
        Speech naturalness & -1.52 & 0.089 & -1.67 \\
        Interaction naturalness & -1.32 & 0.083 & -1.46 \\
        Scenario plausibility & -0.55 & 0.075 & -0.67 \\
        Conv. coherence & -0.50 & 0.087 & -0.64 \\
        \hline
    \end{tabular}
    \caption{Corpus naturalness evaluation. EE: estimated effect (proposed $-$ natural), 1--5 scale; SE: standard error; LB: one sided 95\% lower bound ($\text{EE} - 1.645\,\text{SE}$). Non-inferiority requires LB above the margin of $-0.30$.}
    \label{tab:iw-corpus_naturalness}
\end{table}

\subsection{Intelligent Wakeup Model}

The baseline model (\texttt{gpt-realtime-mini}) answered 158/158 direct and 140/141 contextual turns, but stayed silent for only 18.6\% of non-addressed speech - 516.6 false accepts per hour over 5.28 h. Invocation was not difficult for the baseline; suppression was, and the silence rate ranged from 53\% on procedural narration to 0\% on open-ended debate.

Table~\ref{tab:iw-model_results} shows every trained configuration across different encoders. Neither capacity nor encoder family is the constraint. The one clear separation is pre-trained objective - the AudioSet-tagged Zipformer fell to 0.797 F1 and 0.823 direct recall, indicating that ``what words were said'' transfers to this task and ``what kind of sound it is'' does not. Direct recall reached 0.99 for several encoders while contextual recall never exceeded 0.794, and the two traded against each other. Wake-word turns are effectively solved; the remaining error is concentrated in follow-ups that carry no keyword and must be inferred from context. 

\begin{table}[!t]
    \centering
    \small
    \begin{tabular}{l r c c c}
        \hline
        \textbf{Encoder} & \textbf{Total (M)} &
        \textbf{Direct R} & \textbf{Ctx R} & \textbf{F1} \\
        \hline
        gpt-realtime-mini & -- & 1.000 & 0.993 & 0.179 \\
        whisper-base & 24.0 & 0.943 & 0.794 & 0.894 \\
        whisper-base (MM) & 24.1 & 0.994 & 0.702 & 0.886 \\
        whisper-base (6L) & 31.6 & 0.968 & 0.752 & 0.889 \\
        whisper-small & 91.6 & 0.956 & 0.787 & 0.885 \\
        zipformer-AudioSet & 25.3 & 0.823 & 0.695 & 0.797 \\
        zipformer-small & 25.3 & 0.981 & 0.716 & 0.889 \\
        hubert-base-ls960 & 97.8 & 0.994 & 0.738 & 0.871 \\
        \hline
    \end{tabular}
    \caption{Intelligent wakeup performance across speech encoders.}
    \label{tab:iw-model_results}
\end{table}

\section{Conclusion}

We introduced an intelligent wakeup framework that extends traditional wake-word detection with contextual activation, allowing a voice assistant to remain engaged without requiring repeated explicit invocation. To support this task, we developed a controllable synthetic conversation generation pipeline and released a dataset together with trained models. Experiments show that compact speech encoders can achieve strong direct and contextual wake-up performance, while the remaining challenge lies in inferring implicit follow-up requests from conversational context. Our naturalness evaluation further shows that synthetic scenarios can be plausible and coherent, although a substantial gap to real recorded speech remains. Future work should therefore focus on improving conversation and acoustic realism and on extending contextual wake-up to more diverse settings.

\section{Compliance with Ethical Standards}
Participants were voluntarily recruited through Prolific and compensated. This minimal-risk perceptual study collected no sensitive personal data and required no formal ethical approval.

\section{Acknowledgments}
This work was supported by TCL Research Europe. The authors declare no other relevant financial or nonfinancial interests.

\bibliographystyle{IEEEbib}
\bibliography{bibliography}

\end{document}